%% file: main.tex
\documentclass[sigconf]{aamas}  

\usepackage{booktabs}

\usepackage{subcaption}
\usepackage{textgreek}
\usepackage{multirow}
\usepackage{hhline}
\usepackage{adjustbox}
\usepackage{balance}  
\usepackage{amsmath,amssymb}
\DeclareMathOperator{\E}{\mathbb{E}}

\setcopyright{rightsretained}  
\acmDOI{doi}  
\acmISBN{}  
\acmConference[ALA'18]{Adaptive Learning Agents Workshop (ALA)}{July 2018}{Stockholm, Sweden}  
\acmYear{2018}  
\copyrightyear{2018}  
\acmPrice{}  

\renewcommand\footnotetextcopyrightpermission[1]{} 
\begin{document}

\title{Pre-training Neural Networks with Human Demonstrations for Deep Reinforcement Learning}  

\author{Gabriel V.~de la Cruz, Jr.}
\orcid{0000-0001-8062-5593}
\affiliation{%
 \institution{Washington State University}
 \streetaddress{P.O. Box 642752}
 \city{Pullman} 
 \state{Washington} 
 \postcode{99164-2752}
}
\email{gabriel.delacruz@wsu.edu}

\author{Yunshu Du}
\affiliation{%
 \institution{Washington State University}
 \streetaddress{P.O. Box 642752}
 \city{Pullman} 
 \state{Washington} 
 \postcode{99164-2752}
}
\email{yunshu.du@wsu.edu}

\author{Matthew E. Taylor}
\affiliation{%
 \institution{Washington State University}
 \streetaddress{P.O. Box 642752}
 \city{Pullman} 
 \state{Washington} 
 \postcode{99164-2752}
}
\email{taylorm@eecs.wsu.edu}

\begin{abstract}  
Deep reinforcement learning (deep RL) has achieved superior performance in complex sequential tasks by using a deep neural network as its function approximator and by learning directly from raw images. A drawback of using raw images is that deep RL must learn the state feature representation from the raw images in addition to learning a policy. 
As a result, deep RL often requires a prohibitively large amount of training time and data to reach reasonable performance, making it inapplicable in real-world settings, particularly when data is expensive. In this work, we speed up training by addressing half of what deep RL is trying to solve --- feature learning. We show that using a small set of non-expert human demonstrations during a supervised pre-training stage allows significant improvements in training times.
We empirically evaluate our approach using the deep Q-network and the asynchronous advantage actor-critic algorithms in the Atari 2600 games of Pong, Freeway, and Beamrider. Our results show that pre-training a deep RL network provides a significant improvement in training time, even when pre-training from a small number of noisy demonstrations.
\end{abstract}

%


\maketitle


\input{body}


\newpage
\balance
\bibliographystyle{ACM-Reference-Format}  
\bibliography{main}  

\end{document}

%% file: body.tex
\section{Introduction}

The recent resurgence of neural networks in reinforcement learning (RL) can be attributed to the widespread success of Deep Reinforcement Learning (deep RL), which uses deep neural networks for function approximation \cite{mnih2015human,mnih2016asynchronous}. 
One of the most impressive accomplishments of deep RL is its ability to learn directly from raw images, achieving state-of-the-art results. However, in order to bring the success of deep RL from virtual environments to real-world applications, we must address the lengthy training time that is required to learn a policy.

Deep RL suffers from poor initial performance like classic RL algorithms since it learns \emph{tabula rasa} \cite{sutton1998reinforcement}. In addition, deep RL inherently takes longer to learn because besides learning a policy it also learns a state directly from raw images --- instead of using hand-engineered features, deep RL needs to learn to construct relevant high-level features from raw images.
These problems are consequential in real-world applications with expensive data, such as in robotics, finance, or medicine.

Leveraging humans to provide demonstrations is one method to speed up deep RL. Using human demonstrations in RL is not new \cite{argall2009survey} but only recently has this area gained traction as a possible way of speeding up deep RL \cite{kurin2017atari,vinyals2017starcraft,hester2018learning}.

Speeding up deep reinforcement learning can be achieved by addressing the two problems it is trying to tackle: 1) feature learning and 2) policy learning. In this work, we will focus only on addressing the problem of feature learning by pre-training to learn the underlying features in the hidden layers of the network. We show that by learning better features, an RL agent can achieve better performance without changing its policy learning strategies thus addressing the importance and usefulness of learning good features in an RL problem.
We apply a common technique to deep RL that is widely used to speed up training in deep learning: pre-training a network \cite{erhan2009difficulty,Erhan:2010:WUP:1756006.1756025,yosinski2014transferable}. However, the success of this technique in deep supervised learning is attributed to the large datasets that are available and used to pre-train networks. In deep RL, data are often unavailable or difficult to collect. 

In this work, we propose an approach to speed up deep reinforcement learning algorithms using only a relatively small amount of non-expert human demonstrations. This approach starts by pre-training a deep neural network using human demonstrations through supervised learning. Similar work has shown that this step would learn to imitate the human demonstrator \cite{argall2009survey}. However, in this context, pre-training through supervised learning would implicitly learn the underlying features.

We test our approach with two popular deep RL algorithms, the Deep Q-network (DQN) and the Asynchronous Advantage Actor-Critic (A3C), and evaluate its performance in the Atari 2600 games of Pong, Freeway, and Beamrider \cite{bellemare13arcade}. Our results show speed ups in five of the six cases. The improvements in Pong and Freeway were quite large in DQN, and A3C's improvement on Pong was especially large. The generality of this approach means that it can be easily incorporated into multiple deep RL algorithms.

\section{Related Work}
Our work is not precisely transfer learning, but it is similar to one of the existing transfer learning methods in deep learning. In training deep neural networks for image classification, Yosinski et al.~\shortcite{yosinski2014transferable} have shown how transferring the features learned from existing models allow new models to learn faster, particularly when the datasets are similar. 
In this work, we use a deep learning classifier as the source network to initialize the RL agent's network. 

Existing work on pre-training in RL has shown learning can be improved \cite{abtahi2011deep,anderson2015faster}.
However, their networks have a much smaller number of parameters and state dynamics of the domain are used as network input. In our approach, we use the raw images of the domain as network input and also the RL agent needs to learn the latent features while learning its policy.

Our approach of using supervised learning for pre-training is also similar in spirit to that of Anderson et al.~\shortcite{anderson2015faster}. They pre-train by learning to predict the state dynamics. We instead pre-train by using the game's image frames from the human demo as training data, which are individually labeled by the action taken by the human demonstrator. This setup is similar to how one could derive a policy when learning from demonstration \cite{argall2009survey}.

Another approach to pre-training is to learn the latent features using unsupervised learning through Deep Belief Networks \cite{abtahi2011deep}. Although this pre-training approach differs --- it falls under a different machine learning paradigm --- its goals are similar to our approach in that pre-trained networks learn better than randomly initialized networks.

There are more recent works that leverage humans in deep RL. Christiano et al.~\shortcite{christiano2017deep} use human feedback to learn a reward function. Hester et al.~\shortcite{hester2018learning} similarly pre-train the network with human demonstrations in DQN. However, their pre-training combines the large margin supervised loss and the temporal difference loss, which tries to closely imitate the demonstrator. Our work differs in that only the cross-entropy loss is used 
and we focus on the implicitly learned features. 

The work of Silver et al.~\shortcite{silver2016mastering} trained human demonstrations in supervised learning and used the supervised learner's network to initialize RL's policy network. They tested this approach in a single domain and a huge amount of expert demonstration data was used to train the supervised learner. 
Our work will be the first to provide a comparative analysis as to how this approach impacts deep RL algorithms and how well this approach can complement existing deep RL algorithms when human demonstrations are available. Silver et al.~\shortcite{silver2016mastering} also focus on optimizing the policy learned from humans, while our paper focuses on learning the underlying features. Our work shows that: 1) using only a small set of demonstration data is sufficient enough to gain performance improvements, and 2) a supervised learner can still learn important latent features even when demonstrated human data is from non-experts.

\section{Background: Deep Reinforcement Learning}

An RL problem is typically modeled using a Markov decision process, represented by a 5-tuple $\langle S, A, P, R, \gamma \rangle$. An RL agent explores an unknown environment by taking an action $a \in A$. Each action leads the agent to a state $s \in S$. A reward $r \sim R(s,a, s')$ is given based on the action the agent took and the next state $s'$ it reaches. The goal of an RL agent is to learn to maximize the expected return value $R_t = \sum_{k=0}^{\infty} \gamma^k r_{t+k}$ for each state at time $t$. The discount factor $\gamma \in (0,1]$ determines the relative importance of future and immediate rewards.

\subsection{Deep Q-network}
The first successful deep RL method, deep Q-network (DQN), learns to play 49 Atari games directly from screen pixels by combining Q-learning with a deep convolutional neural network \cite{mnih2015human}. In shallow Q-learning, an agent learns a state-action value function $Q^*(s, a) = \E_{s'}[r+\gamma \max_{a'}Q^* (s', a')|s, a]$, which is the expected discounted reward determined by performing action $a$ in state $s$ and thereafter performing optimally \cite{watkins1992q}. The optimal policy $\pi^*$ can be deduced by following actions that have the maximum Q value, $\pi^* = argmax_{a}Q^*(s,a)$.

Directly computing the Q value is not feasible when the state space is large or continuous (e.g., in Atari games). The DQN algorithm uses a convolutional neural network as a function approximator to estimate the Q function $Q(s, a; \theta) \approx Q^*(s, a)$, where $\theta$ is the network's weight parameters. For each iteration $i$, DQN is trained to minimize the mean-squared error (MSE) between the Q-network and its target $y = r + \gamma max_{a'}Q(s', a';\theta_{i}^-)$, where $\theta_{i}^-$ is the weight parameters for the target network that was generated from previous iterations. All rewards are clipped to 1 when positive, -1 when negative, and 0 when unchanged. 
The loss function at iteration $i$ can be expressed as $L_{i}(\theta_i) = \E_{s, a, r, s'} [(y - Q(s, a; \theta_i))^2]$, where $\{s, a, r, s'\}$ are state-action samples drawn from experience replay memory with a minibatch of size 32. The use of a target network, reward clipping, and an experience replay memory are essential to stabilize learning. In addition, the $\epsilon$-greedy policy is used by the agent to obtain sufficient exploration of the state space.

\subsection{Asynchronous Advantage Actor-critic}

There are a few drawbacks of using experience replay memory in the DQN algorithm. First, storing all experiences is space-consuming and could slow down learning. Second, using replay memory limits DQN to off-policy algorithms. The asynchronous advantage actor-critic (A3C) algorithm was proposed to overcome these problems \cite{mnih2016asynchronous}.  

A3C combines the actor-critic algorithm with deep RL. It differs from value-based algorithms (e.g., Q-learning) where only a value function is learned --- an actor-critic algorithm is policy-based and maintains both a policy function $\pi(a_t|s_t;\theta)$ and a value function $V(s_t;\theta_v)$ \cite{sutton1998reinforcement}. The policy function is called the actor, which takes actions based on the current policy $\pi$. The value function is called the critic, which serves as a baseline to evaluate the quality of the action by returning the state value $V^\pi(s_t;\theta_t)$ for the current state under policy $\pi$. The policy is directly parameterized and improved via policy-gradient. To reduce the variance in policy gradient, an advantage function is used and calculated as $A(a_t, s_t;\theta,\theta_v) = \sum_{i=0}^{k-1} \gamma^{i}t_{t+i} + \gamma^{k}V(s_{t+k};
\theta_v) - V(s_t;\theta_{v})$ at time step $t$ for action $a_t$ at state $s_t$, where $k$ is upper-bounded by $n$, the number of steps used for n-step return update. The loss function for A3C is $L(\theta) = \nabla_\theta \log \pi(a_t|s_t;\theta)A(a_t, s_t;\theta,\theta_v)$.

In A3C, $k$ actor-learners are running in parallel with their own copies of the environment and the parameters for the policy and value function. This enables exploration of different parts of the environment and therefore observations will not be correlated. This mimics the function of experience replay memory in DQN while being more efficient in space and training time. Each actor-learner performs a parameter update every $t_{max}$ actions, or when a terminal state is reached --- this is similar to using mini-batches, as is done in DQN. Updates are synchronized to a master learner that maintains a central policy and value function, which will be the final policy upon the completion of training.

\section{Pre-Training Networks for Deep RL}

Deep reinforcement learning can be divided into two sub-tasks: feature learning and policy learning. Deep RL in itself has already succeeded in learning both tasks simultaneously. However, learning both tasks also makes learning in deep RL very slow. We believe that by addressing the feature learning task, deep RL agents can better focus on learning the policy. We learn the features by pre-training the network using human demonstrations from non-experts. We assume here that humans provide correct labels through actions demonstrated while playing the game. We will refer to our approach as the \emph{pre-trained model}.

We first apply the pre-trained model approach in DQN and refer to it as the \emph{pre-trained model for DQN (PMfDQN)}. In PMfDQN, we train a multiclass-classification deep neural network with a softmax cross entropy loss function. The loss is minimized using the Adam optimizer \cite{kingma2014adam} with the following hyperparameters: step size $\alpha=0.0001$, stability constant $\epsilon=0.001$, and Tensorflow's default exponential decay rates $\beta$. The network architecture for the classification follows the same structure of the hidden layers of DQN, which has three convolutional layers (\emph{conv1}--\emph{conv3}) and one fully connected layer (\emph{fc1}) \cite{mnih2015human}. The classifier's output layer has a single output for each valid action and is trained using the cross-entropy loss instead of the TD loss. The learned weights and biases from the classification model's hidden layers are used to initialize the DQN network, instead of random initialization. When using all layers of the pre-trained model (including the output layer), normalization of the parameters of the output layer was necessary to achieve a positive result. To normalize the output layer, we keep track of the maximum value of the output layer during training, which is used as a divisor to all the weights and biases during initialization with the pre-trained model. Without normalization, the values of the output layer tend to explode. We also load the human demonstrations in the replay memory, thus removing the need for DQN to take uniform random actions for 50,000 frames to initially populate the replay memory \cite{mnih2015human}. 

The pre-trained model method can also be applied in A3C, which we will refer to as the \emph{pre-trained model for A3C (PMfA3C)}. In PMfA3C, we pre-train the multiclass-classifier using the same hyperparameters and optimization method as mentioned in PMfDQN while experimenting with two different network structures. The first network uses a structure with three convolutional layers (\emph{conv1}--\emph{conv3}) and one fully connected layer (\emph{fc1}), but without the long short-term memory (LSTM) cells \cite{sharma2017learning}. The output layer is the same as in PMfDQN. The second network is inspired by one-vs.-all multiclass-classification and multitask learning \cite{caruana1998multitask}. It differs from the first network as it uses multiple heads of output layers where each class or action has its own output layer. Each individual output layer becomes a one-vs.-all classification. During each training iteration, a uniform probability distribution is used to select which output layer to train. In each iteration, gradients are backpropagated to the shared hidden layers of the network. In both multiclass networks, only the hidden layers are used to initialize A3C's network.

Since DQN uses experience replay memory \cite{lin1992self}, it is also possible to pre-train just by loading the human demonstrations in the replay memory. We refer to this experiment as \emph{pre-training in DQN (PiDQN)}. While somewhat naive, this is still an interesting method as it allows the DQN agent to learn both the features and policy without any interaction with the actual Atari environment. However, this pre-training method does not generalize to A3C and/or other deep RL algorithms that do not use a replay memory. We would like to address this in future work by applying this naive approach to an alternative version of A3C that uses a replay memory \cite{wang2016sample}.

Lastly, we conducted additional experiments in DQN that combines PMfDQN and PiDQN, with the goal of exploring whether a combined approach would achieve a greater performance in DQN.

\section{Experimental Design}
\label{experiments}

\begin{figure}[H]
    \centering
     \includegraphics[width=0.99\columnwidth]{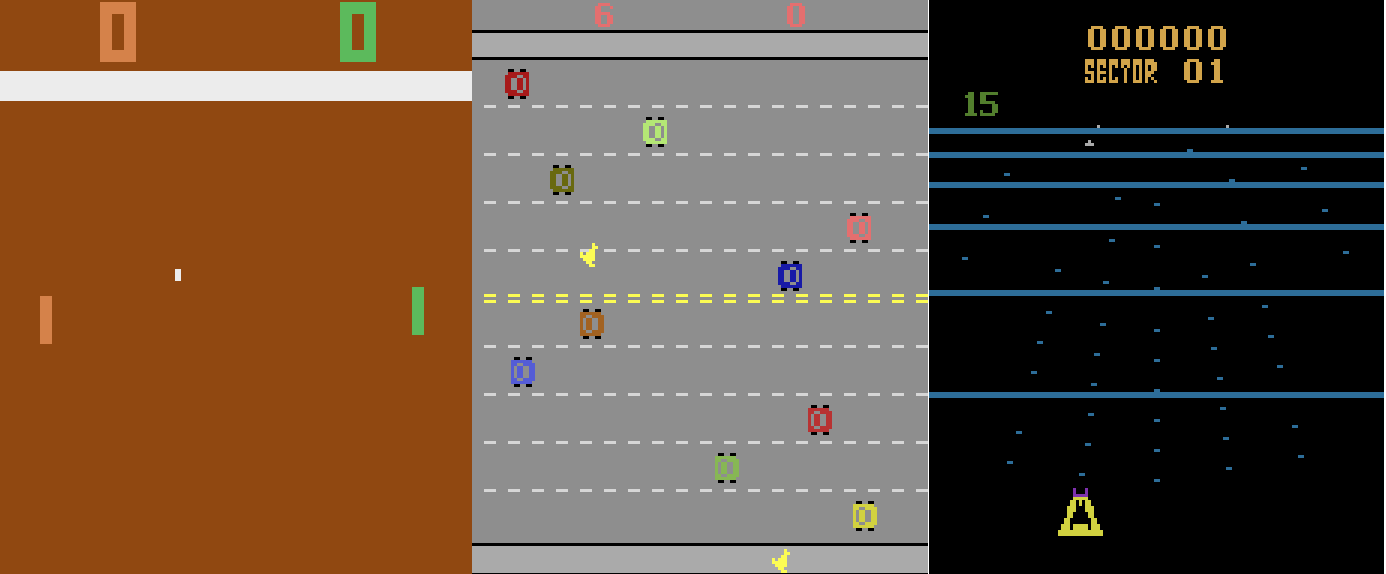} 
     \caption{Atari 2600 game screenshot of Pong, Freeway and Beamrider, from left to right.} 
     \label{fig:atari_games}
\end{figure}

We test our approach in three Atari games: Pong, Freeway, and Beamrider, as shown in Figure~\ref{fig:atari_games}.
The games have 6, 3, and 9 actions, respectively. We use OpenAI Gym's deterministic version of the Atari 2600 environment with an action repeat of four \cite{openaigym}.

\begin{table*}[t!]
\caption{Summary of pre-training experiments.}
\label{table:experiments}
\centering
\begin{adjustbox}{width=0.85\textwidth}
\begin{tabular}{l|c}
Method      & Summary  \\ \hline \hline
PiDQN                             & pre-train in DQN for 150,000 iterations, batch size of 32 \\ \hline
PMfDQN                            & initialize DQN with pre-trained model, pre-train for 150,000 iterations, batch size of 32 \\ \hline
PMfDQN+PiDQN                      & initialize DQN with pre-trained model and continue to pre-train in DQN \\ \hline
PMfDQN+PiDQN ($\epsilon=0.1$)     & low initial exploration rate \\ \hline
PMfDQN (random demo)              & pre-train model with random demonstrations  \\ \hline
PMfDQN (no fc2)                   & initialize with pre-trained model excluding output layer \\ \hline
PMfA3C                            & initialize A3C with pre-trained model, pre-train for 150,000 iterations, batch size 32    \\ \hline
PMfA3C (1-vs-all)                 & pre-train model using one-vs-all multi-class classification, longer pre-training \\ \hline
PMfA3C (1-vs-all, 1-demo)         & pre-train model using only one out of the five demonstrated game play \\ \hline \hline
\end{tabular}
\end{adjustbox}
\end{table*}

We use the same network architecture and hyperparameters for DQN as was done in the original work \cite{mnih2015human}. For the LSTM-variant of A3C, we follow the work of Sharma et al.~\shortcite{sharma2017learning} as their work closely replicates the results of the original A3C algorithm \cite{mnih2016asynchronous}. However, note that there are two key differences from the original A3C work. First, while using the same network architecture with three convolutional layers, the fully connected layer was modified to have 256 units (instead of 512) to connect with the 256 LSTM cells that followed. Second, we use $t_{max}=20$ instead of $t_{max}=5$. We use 16 actor-learner threads for all A3C experiments.

In both DQN and A3C, we use the four most recent game frames as input to the network where each frame is pre-processed. We also use the same evaluation technique for both DQN and A3C such that the average reward over 125,000 steps was recorded. In addition, DQN is evaluated using a $\epsilon$-greedy action selection method, where $\epsilon=.05$. In A3C, it is evaluated as a stochastic policy where it uses the output policy as action probabilities.

\subsection{Collection of Human Demonstration}
We are using OpenAI Gym's keyboard interface to allow a human demonstrator to interact with the Atari environment. The demonstrator is provided with game rules and a set of valid actions with their corresponding keyboard keys for each game. The action repeat is set to one to provide smoother transitions of the games during human play, whereas the action repeat is set to four during training \cite{mnih2015human,mnih2016asynchronous}.
During the demonstration, we collect every fourth frame of the game, saving the game state using the game's image, action taken, reward received, and if the game's current state is a terminal state. The format of the stored data follows the structure of the experience replay memory used in DQN.

The non-expert human demonstrator plays five rounds for each game. Each round has a maximum of five minutes of playing time. The demonstration ends when the game reaches the time limit or when game terminates --- whichever comes first.
Table~\ref{table:human_demo} provides a breakdown of human demonstration size for each game and the human performance level.

\section{Results}

This section presents and discusses results from pre-training deep RL's network for DQN and A3C.

\begin{table}[pt!]
\caption{Human demonstration over five plays per game.}
\label{table:human_demo}
\centering
\begin{tabular}{l|c|c|c}
Game      & Worst Score & Best Score & \# of Frames \\ \hline \hline
Beamrider & 2,160       & 3,406      & 11,205  \\ \hline
Freeway   & 28          & 31         & 10,241 \\ \hline
Pong      & -10         & 5          & 11,265 \\ \hline \hline
\end{tabular}
\end{table}

\subsection{DQN}

\begin{figure}[pt!]
  \centering
    \begin{subfigure}[h]{0.99\columnwidth}
        \includegraphics[width=\columnwidth]{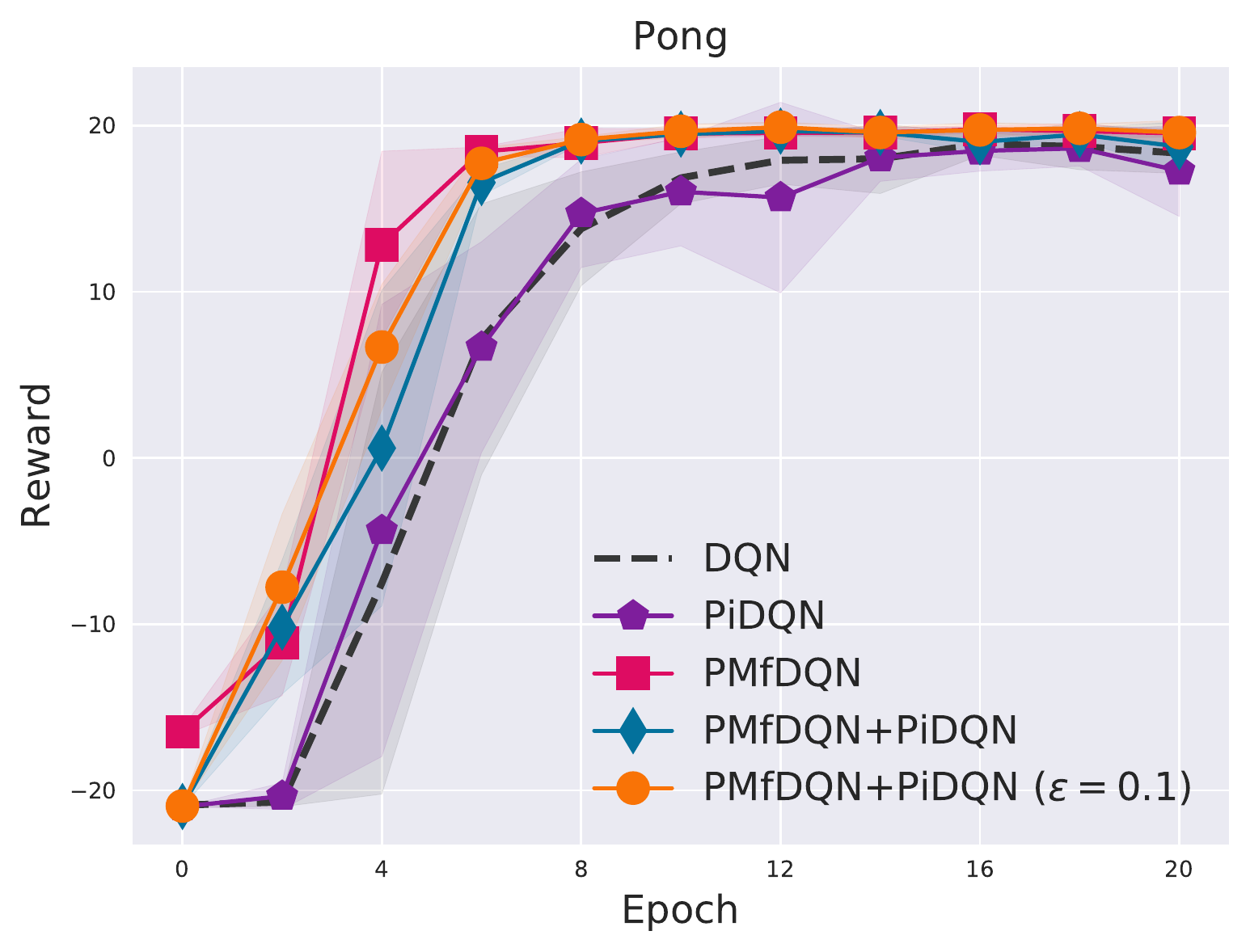}
    \end{subfigure}
    
    \begin{subfigure}[h]{0.99\columnwidth}
        \includegraphics[width=\columnwidth]{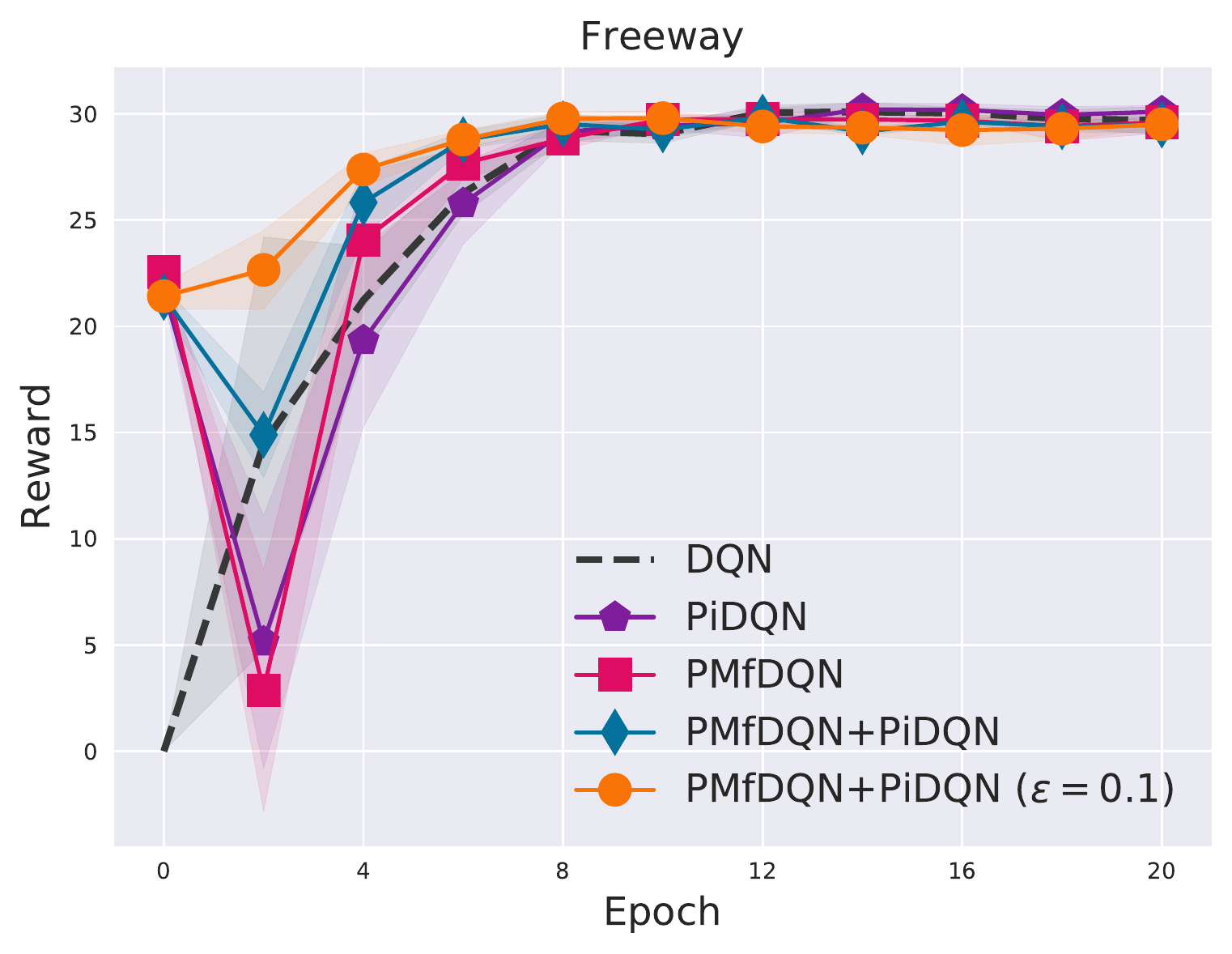}
    \end{subfigure}
    
    \begin{subfigure}[h]{0.99\columnwidth}
        \includegraphics[width=\columnwidth]{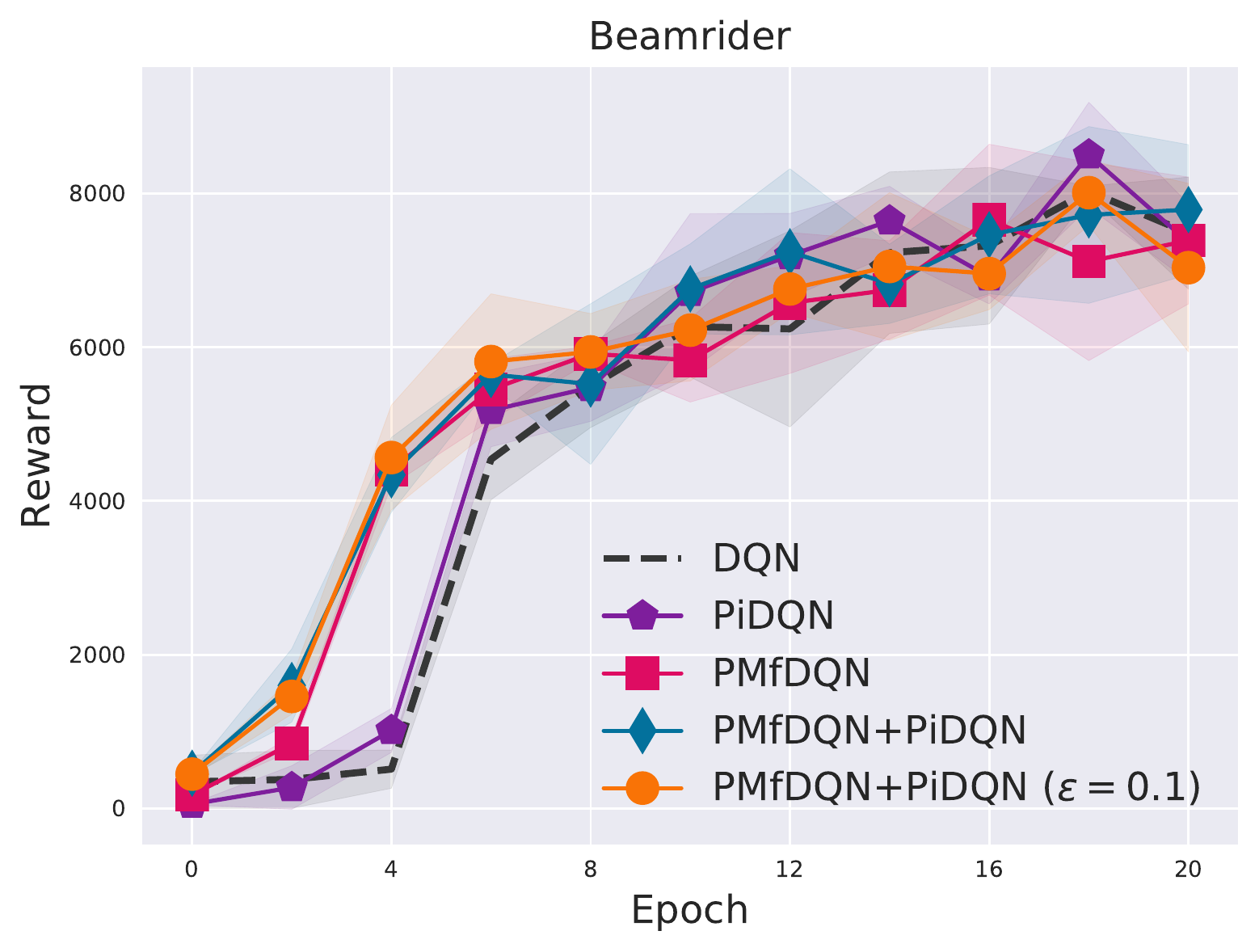}
    \end{subfigure}
    \caption{Performance evaluation of the ablation studies for Pong using DQN. The results are the average testing score over four trials where the shaded regions correspond to the standard deviation.}
    \label{fig:dqn_result}
\end{figure}

Using PMfDQN, we trained one multiclass-classification network for each Atari game with the human demonstration dataset. Each training run was conducted using a batch size of 32 for 150,000 training iterations. The number of training iterations was determined to be the shortest number of iterations where the training loss for all games converges approximately to zero. The trained classifiers provided us the pre-trained models which were then used to initialize DQN's network weights and biases. 
Figure~\ref{fig:dqn_result} shows the results that PMfDQN speeds up training in all three Atari games. We also tested PiDQN with the same number of pre-training iterations as in PMfDQN. PiDQN for Beamrider has shown some performance improvement with an average total reward of 5,120 compared to DQN's average total reward of 4,894. However, PiDQN in Pong and Freeway either follows a similar learning trajectory as DQN or slightly worse.
Our experiments show although naive pre-training could provide speedups under some cases, supervised pre-training is the essential component when using demonstrations.

To see if we can further improve DQN through pre-training, we used PMfDQN followed by PiDQN with 150,000 pre-training iterations each. Figure~\ref{fig:dqn_result} shows the result for the combined method (PMfDQN+PiDQN). This method showed a slight improvement in playing Freeway but less improvement in Pong and Beamrider when compared to only using PMfDQN. We found these results surprising since we hypothesized that more improvement should be expected with more pre-training. 
One possible reason for this could be due to the high initial exploration rate $\epsilon=1$ in DQN at the beginning of training. Under this setting, the agent would be taking entirely random actions until the value of $\epsilon$ has decayed to a much lower exploration rate. In the original DQN, $\epsilon$ is decayed over one million steps, resulting in a replay memory with a good amount of experiences executed through random action \cite{mnih2015human}. This may have an adverse effect on what has already been learned from the pre-training steps. Therefore, we instead initialized $\epsilon=0.1$ when using the PMfDQN+PiDQN combined method.
Results for the combined method as shown in Figure~\ref{fig:dqn_result} revealed that combining PMfDQN with PiDQN using a low initial exploration rate was comparable to PMfDQN by itself and was even better for Freeway. DQN's high initial exploration rate can be detrimental in costly real-world applications. By using a small amount of human demonstration, we can now minimize exploration without affecting agent's overall learning performance.

To measure improvement for each pre-training method, we computed the average total reward for each trial and compared each pre-training method against the DQN baseline. In PMfDQN, the average total reward was higher than the baseline in all three games, although it was only statistically significant for Pong and Freeway.
However, the improvements in both PMfDQN+PiDQN and PMfDQN+PiDQN ($\epsilon=0.1$) were statistically significant in all three games as indicated by t-test with $p<.05$.

\subsubsection{Ablation Studies}

\begin{figure}[pt!]
    \includegraphics[width=0.99\columnwidth]{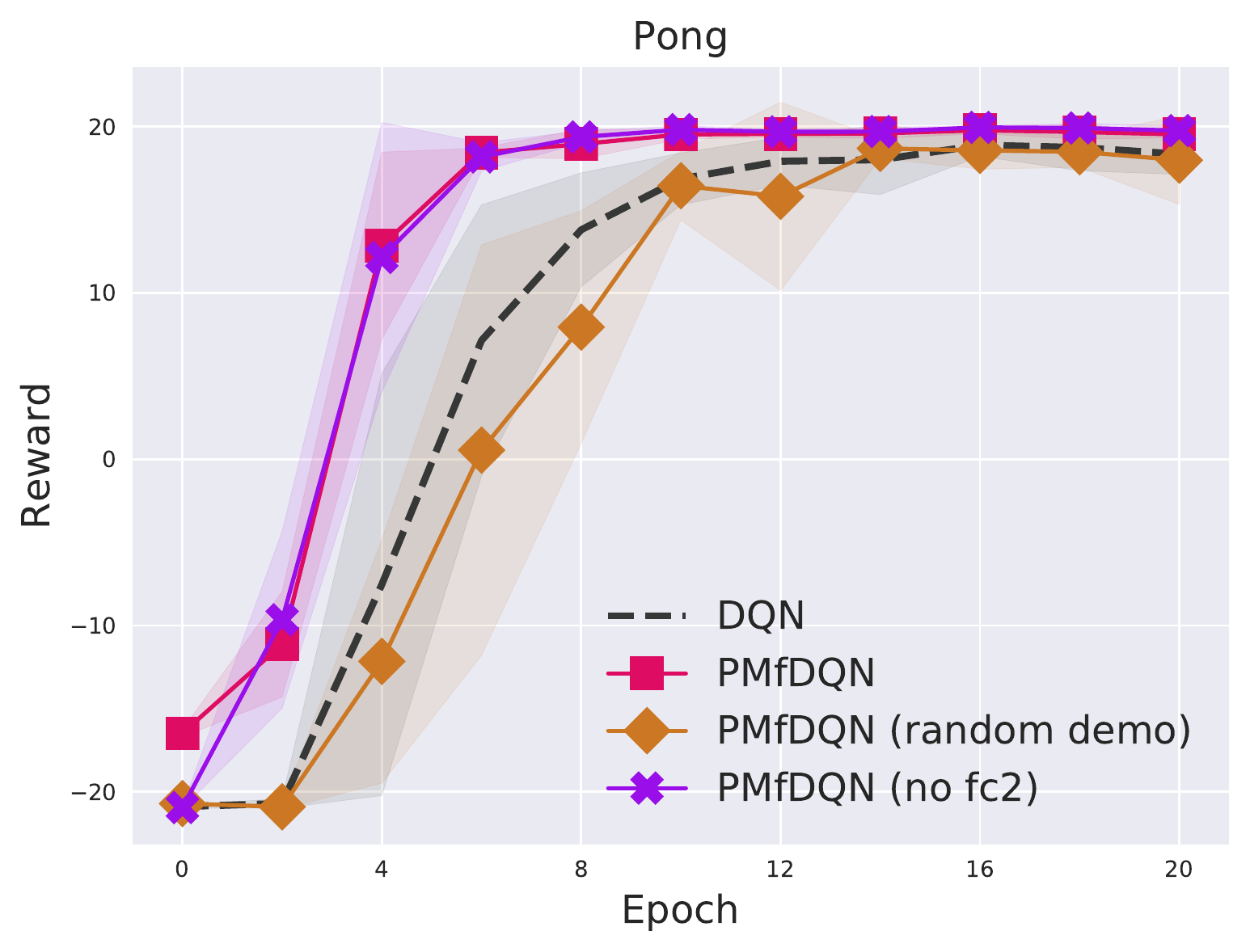}
    \caption{Performance evaluation on the ablation studies for Pong using DQN. The results are the average testing score over four trials where the shaded regions correspond to the standard deviation.}
    \label{fig:dqn_pong_ablation}
\end{figure}

We consider two modifications to PMfDQN to further analyze its performance. In our first ablation study, we replaced human demonstrations with random demonstrations. We were interested in knowing how important it is to use human demonstrations in comparison with using a random agent. We conducted this experiment in Pong and the results in Figure~\ref{fig:dqn_pong_ablation} showed that pre-training with random demonstrations was worse than the DQN baseline. This experiment indicated that there was a need for some level of competency from the demonstrator in order to extract useful features during pre-training. However, even with worse results, our approach appears to still converge to a policy similar to baseline DQN.

In our second ablation study, we excluded the second fully connected layer (fc2) (i.e., the output layer) when initializing the DQN network with the pre-trained model. This will allow us to know if supervised learning does learn important features, particularly in the hidden layers. 
Empirically, when excluding the output layer, results in Figure~\ref{fig:dqn_pong_ablation} showed that even though the initial jumpstart was lost, the training time to reach convergence is not different from the time when using all layers. This indicated that it was actually the features in the hidden layers that provided most of the improvement in the training speed. This was not surprising since the output layer of a classifier was trying to learn to predict what action to take given a state without any consideration for maximizing the reward. Additionally, when learning from only a small amount of data where human performance was relatively poor (Table~\ref{table:human_demo}), the classifier's policy would be far from optimal.

\subsection{A3C}
\begin{figure}[pt!]
  \centering
    \begin{subfigure}[h]{0.99\columnwidth}
        \includegraphics[width=\columnwidth]{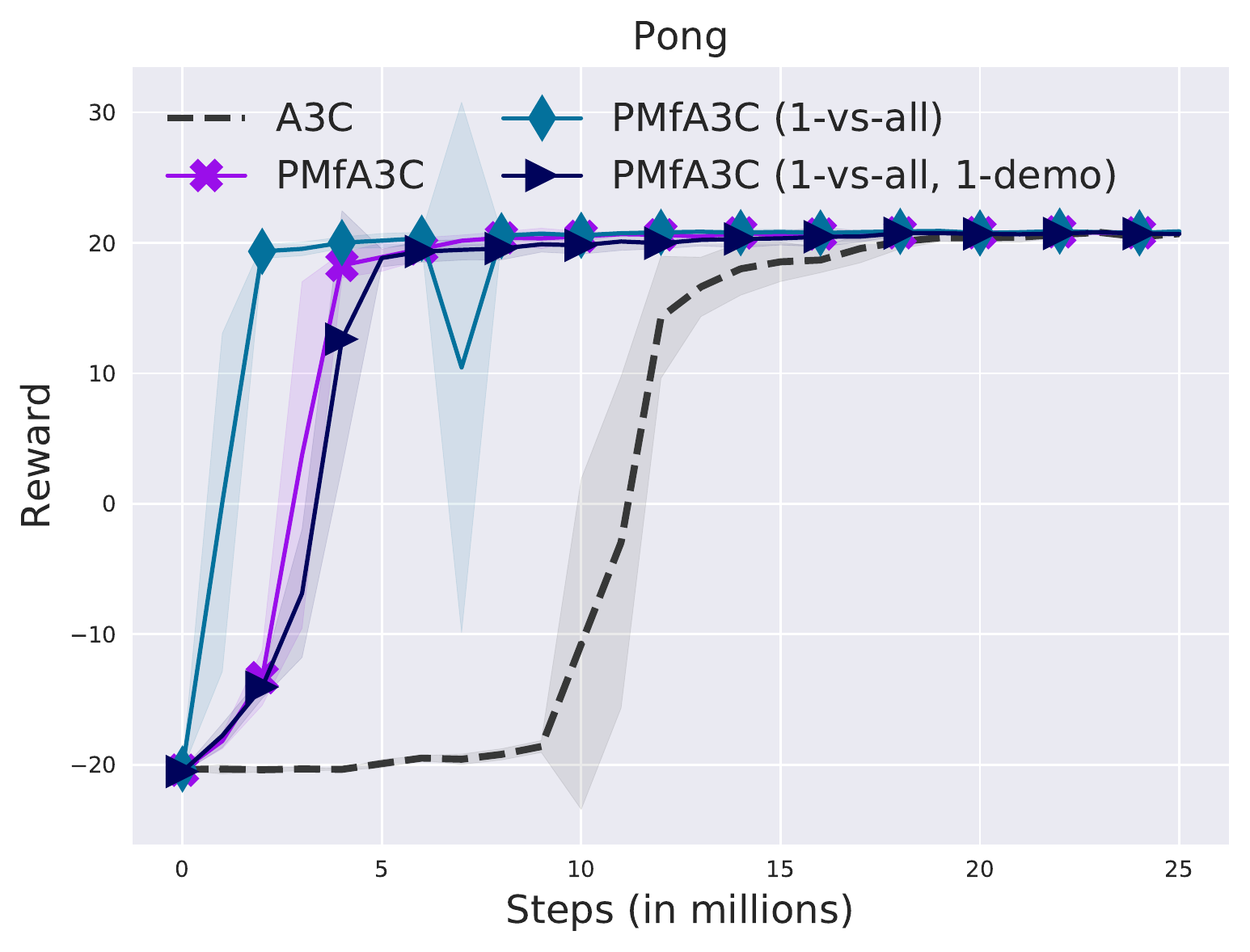}
    \end{subfigure}
    
    \begin{subfigure}[h]{0.99\columnwidth}
        \includegraphics[width=\columnwidth]{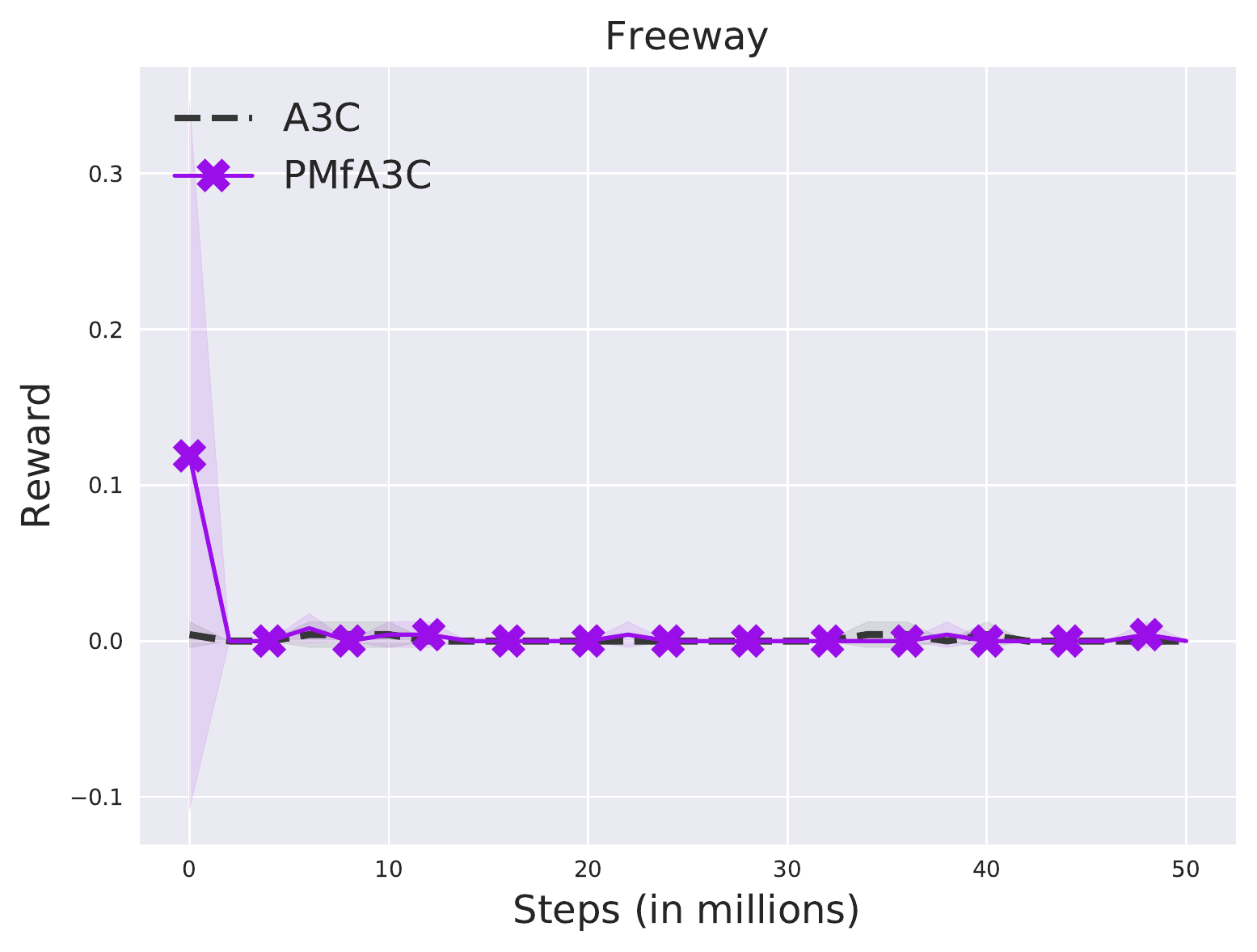}
    \end{subfigure}
    
    \begin{subfigure}[h]{0.99\columnwidth}
        \includegraphics[width=\columnwidth]{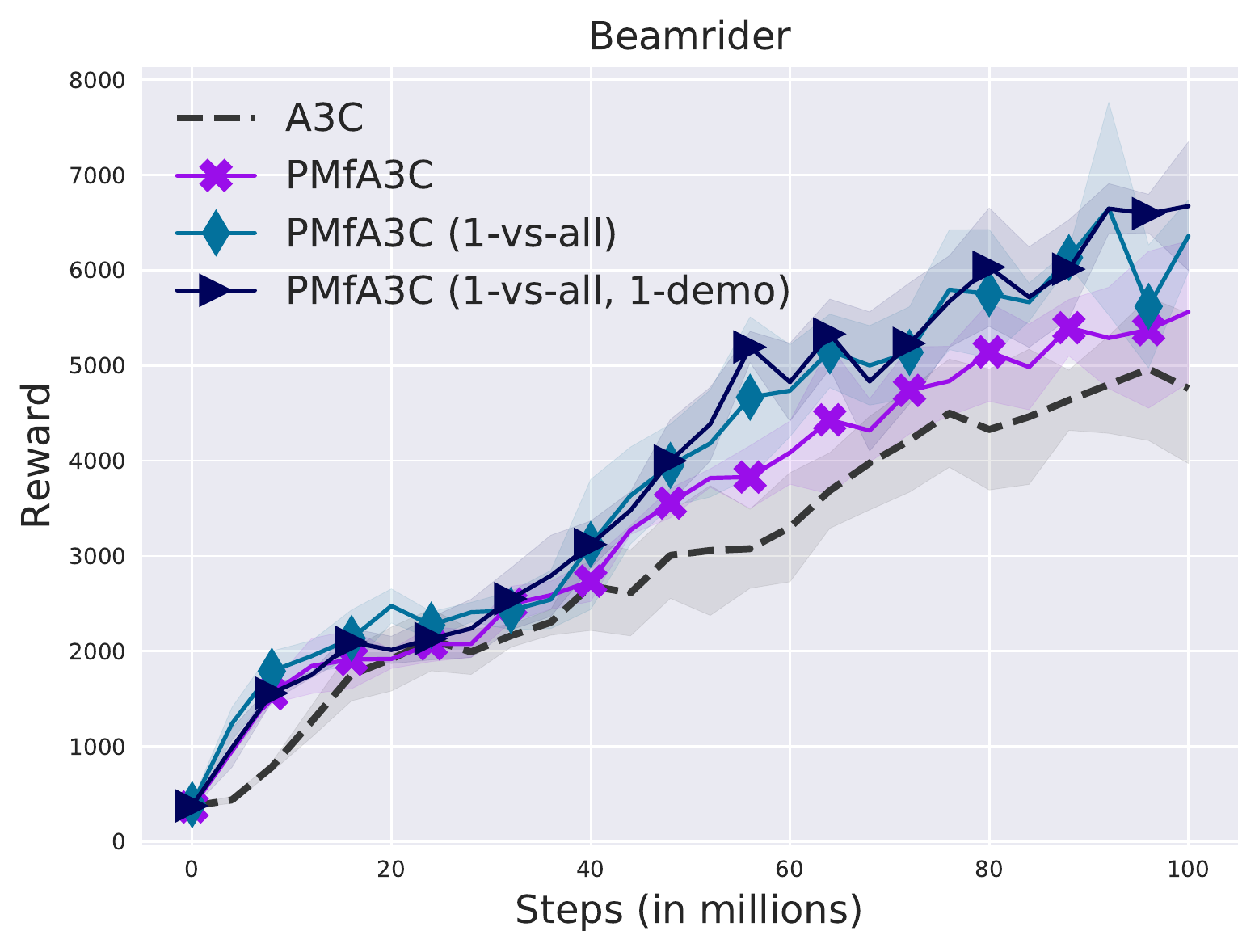}
    \end{subfigure}
    
    \caption{Performance of baseline and pre-training using A3C. The x-axis is the number of training steps which is also the number of visited game frames among all parallel workers with frame skip. The y-axis is the average testing score over four trials where the shaded regions correspond to the standard deviation. Note that all PMfA3C experiments do not use the output layer from the pre-trained model.}
    \label{fig:a3c_result}
\end{figure}

Using PMfA3C, we also pre-trained multiclass-classification networks for each Atari game with human demonstrations, similar to what was done in PMfDQN with a batch of 32 for 150,000 training iterations. Since the network for the LSTM-variant of A3C used LSTM cells with two output layers, we only initialize A3C's network with the pre-trained model's hidden layers. In Figure~\ref{fig:a3c_result}, results showed improvements in the training time in both Pong and Beamrider, with a much higher improvement in Pong. 

However, there was no improvement in Freeway. This was within our expectation since the baseline performance of Freeway was poor in the original A3C work \cite{mnih2016asynchronous} (shown in Figure~\ref{fig:a3c_result} baseline). We would like to emphasize that our approach focused on learning features without addressing improvements in policy --- no improvements in Freeway with our approach were expected. Freeway in A3C needs a better way of exploring states in order to learn a near-optimal policy for the game. This is something we will try to address in future work.

With strong improvements observed in A3C, can we still gain further improvements if we pre-train our classification network longer? We then tried longer training using the one-vs.-all multiclass-classification network with shared hidden layers. Since each class or action was trained independently, we can now observe the different convergence of the training loss for each class. This allowed us to use the same technique of training until the training loss for all classes was approximately zero. Using the one-vs.-all classification, we pre-trained for 450,000 iterations in Pong and 650,000 iterations in Beamrider. Training longer resulted in a very large improvement in pong and a slight improvement for Beamrider, as shown in Figure~\ref{fig:a3c_result}.

\begin{figure*}[t!]
  \centering
    \includegraphics[width=0.99\textwidth]{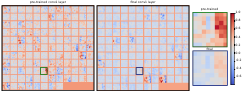}
    \caption{Visualization of the normalized weights on Pong's first convolutional layer using PMfA3C. The weights (filters) are from a pre-trained classification network trained for 150,000 iterations (left image), and from the final weights after 50 million training steps in A3C (right image). To better illustrate the similarity of the weights, we provided two zoomed-in images of a particular filter from pre-trained conv1 (green box) and final conv1 (blue box).}
    \label{fig:weights_visual}
\end{figure*}

The last experiment we conducted was to test whether important features could still be learned even with a much smaller number of demonstrations, in this case, a single round of the game that was only five minutes of demonstration. We used one-vs.-all classification network to pre-train for Pong with only 2,253 game frames with 250,000 training iterations and similarly for Beamrider with 2,232 game frames with 300,000 training iterations. In Figure~\ref{fig:a3c_result}, results for both Pong and Beamrider show improvement with only this small number of demonstrations.
It was even more remarkable in Beamrider, as results were as good as pre-training with the full set of the human demonstrations.

\subsection{Additional Analysis}
In order to understand what was accomplished with pre-training, we looked closer at the weights of the network layers (i.e., filters) to determine how much pre-trained features contributed to the final features learned. Thus, we further investigated how similar the initial weights $\hat{\theta}$ of a deep RL network were to its final weights $\theta$ for each layer after learning a near-optimal policy. We can quantify the similarity by finding the difference between the weights using the mean squared error $\text{MSE}=\frac{1}{n} \sum_{i=1}^{n} (\hat{\theta}_i - \theta_i)^2 $. Smaller MSE indicates a higher similarity between layers. Table~\ref{table:mse_similarity} shows that there was a higher similarity in the pre-training approach compared to random weight initialization. Furthermore, we looked at the visualization of each hidden layer and observed that the weights learned from classification and used as initial values in deep RL's network provided features that were retained even after training in deep RL as shown in Figure~\ref{fig:weights_visual}.

\begin{table}[t]
\caption{Evaluation of the similarity of features for each hidden layer. The mean squared error (MSE) is computed between the weights from a randomly initialized A3C network (baseline) and the final weights. Similarly, when using a pre-trained model as the initial weights.}
\label{table:mse_similarity}
\centering
\begin{adjustbox}{width=\columnwidth}
\begin{tabular}{l|c|c|c|c}
\multirow{2}{*}{Layer} & \multicolumn{2}{c|}{MSE (Pong)} & \multicolumn{2}{c}{MSE (Beamrider)} \\ \hhline{~----}
       & Baseline                & Pre-train              & Baseline                & Pre-train \\ \hline \hline
conv1  & $1.03\times10^{-2}$ & $3.94\times10^{-3}$ & $3.32\times10^{-2}$ & $2.53\times10^{-2}$ \\ \hline
conv2  & $8.02\times10^{-3}$ & $8.00\times10^{-4}$ & $8.50\times10^{-3}$ & $4.35\times10^{-3}$ \\ \hline
conv3  & $7.13\times10^{-3}$ & $3.26\times10^{-4}$ & $7.11\times10^{-3}$ & $2.39\times10^{-3}$ \\ \hline
fc1    & $9.57\times10^{-4}$ & $7.54\times10^{-5}$ & $1.07\times10^{-3}$ & $3.29\times10^{-4}$ \\ \hline \hline
\end{tabular}
\end{adjustbox}
\end{table}

\section{Discussion and Conclusion}
The pre-training approach worked very well in Pong. This success can be explained by the human demonstration data the classifier was pre-trained with, and the simplicity of Pong environment. Pong's states are highly repetitive when compared to the other game environments that are more dynamic. The Beamrider has the most complex environments among all three games because it has different levels of varying difficulty. Although Freeway's game state is also repetitive, A3C's inability to learn a good policy is a problem that leans more towards policy learning, which is not addressed in our approach.

Human demonstrations are an essential part of the success of our approach. It is important to understand how the demonstrator's performance and the amount of demonstration data affect the benefits of pre-training in future work. Future work will consider using recently released human demonstration datasets for Atari \cite{kurin2017atari} and Starcraft II \cite{vinyals2017starcraft}.

Another issue that needs to be addressed in regards to the human demonstrations is that they suffer from highly imbalanced classes (actions). This is attributed to: 1) sparsity of some actions (e.g., the torpedo action in Beamrider is limited to three uses at each level), 2) actions that are closely related (e.g., in Beamrider, there is a left and right action plus combined actions of left-fire and right-fire --- a demonstrator would usually just use the native actions of left and right action alone and use the fire action by itself), and 3) games having a default no-operation action.

In a previous study, the authors show that the classifier will learn a policy that tends to bias towards the majority classes when the imbalance problem is not addressed \cite{he2009learning}. It is interesting that the classifier is still able to learn important features without handling this issue. We see this as an interesting future work and hope to explore if better features can be learned by handling the class imbalance problem, therefore, leads to further improvements. 

As we investigate further ways to improve our approach, we know there is a limit to how much improvement pre-training can provide without addressing policy learning. In our approach, we have already trained a model with a policy that tries to imitate the human demonstrator thus we can extend this work by using the pre-trained model's policy to provide advice to the agent (e.g., \cite{2017IJCAI-Wang}).

Overall, learning a policy directly from raw images through deep neural networks is a major factor why learning is slow in deep RL. This paper has demonstrated that our method of initializing deep RL's network with a pre-trained model can significantly speed up learning in deep RL.

\begin{acks}

The A3C implementation was a modification of \url{https://github.com/miyosuda/async_deep_reinforce}. The authors would like to thank Sahil Sharma and Kory Matthewson for providing very useful insights on the actor-critic method. We also thank NVidia for donating a graphics card uses in these experiments.

\end{acks}